# Routine Outcome Monitoring in Psychotherapy Treatment using Sentiment-Topic Modelling Approach

Noor Fazilla Abd Yusof#, Chenghua Lin*

# Centre for Advanced Computing Technology (C-ACT), Fakulti Teknologi Maklumat dan Komunikasi (FTMK), Universiti Teknikal Malaysia Melaka, Malaysia,
E-mail: elle@utem.edu.my

* Chenghua Lin, Computer Science Department, University of Sheffield, United Kingdom,
E-mail: c.lin@sheffield.ac.uk

*Abstract*— **Despite the importance of emphasizing the right psychotherapy treatment for an individual patient, assessing the outcome of the therapy session is equally crucial. Evidence showed that continuous monitoring patient's progress can significantly improve the therapy outcomes to an expected change. By monitoring the outcome, the patient's progress can be tracked closely to help clinicians identify patients who are not progressing in the treatment. These monitoring can help the clinician to consider any necessary actions for the patient's treatment as early as possible, e.g., recommend different types of treatment, or adjust the style of approach. Currently, the evaluation system is based on the clinical-rated and self-report questionnaires that measure patients' progress pre- and post-treatment. While outcome monitoring tends to improve the therapy outcomes, however, there are many challenges in the current method, e.g. time and financial burden for administering questionnaires, scoring and analysing the results. Therefore, a computational method for measuring and monitoring patient progress over the course of treatment is needed, in order to enhance the likelihood of positive treatment outcome. Moreover, this computational method could potentially lead to an inexpensive monitoring tool to evaluate patients' progress in clinical care that could be administered by a wider range of health-care professionals.**

*Keywords*— **Psychotherapy treatment; sentiment-topic model, progress monitoring outcome, depression**

## I. INTRODUCTION

Depression is a common illness that has a major contribution to the overall global burden of disease. It can affect one's thinking styles and behaviour, including low energy, loss appetite, reduce concentration and intense emotions of hopelessness and negativity. It has been reported that more than 264 million people of all ages suffer from depression [1]. Considering the high burden of mental illnesses, therefore, treatment recommendations for an individual patient are highly important. The common options for treating depression, i.e. medication, psychotherapy or a combination of both [2]–[4], greatly depend on the individual cases. Psychotherapies such as cognitive-behavioural therapy (CBT) [4]–[6] and interpersonal psychotherapy (IPT) [7], [8] can have substantial effects for treating depression effectively.

In this paper, we tackle the research challenge of monitoring the progress of psychotherapy sessions. Individual psychotherapy counselling transcripts were used to identify the outcome of each therapy session. Evidence [6], [7], [9], [10] show that by regularly identifying the patients who are regressing during the treatment could further improve the therapy outcomes. In contrast, the post-mortem analysis after the patient completed the treatment holds very limited adjustment that can be implied in order to improve the outcomes of the treatment (i.e. no sign of improvement or even worsen conditions) [9].

Commonly in clinical settings, measuring outcomes of patients' therapy sessions are commonly based on the regular self-reports i.e. patient filled out questionnaires at the beginning of treatment, during the treatment, and at the end of a treatment [11]. This progress monitoring known as Routine Outcome Monitoring (ROM) is important in psychotherapy session as it has been used as a tool to assess the patient's progress, to evaluate the treatment, and decide its future course [9]. The main purpose of outcome monitoring is to serve as a threshold to evaluate the progress of a patient, to see whether the ongoing treatments have a positive or negative impact. Several rating scales based on the self-reported questionnaires system have been introduced in the mental health routine practice. These including Outcome Questionnaire System (OQ-45) [10], the Partners for Change System (PCOMS) [12], and the Clinical Outcomes in Routine Evaluation (CORE) [13].

While progressively monitor psychotherapy session is proven to increase the chances of a positive outcome, current monitoring system, however, suffers from few challenges e.g. (i) extra work and effort needed for administering the questionnaires, (ii) time constraints for scoring and analysing the results, (iii) lead to an unnecessary burden on patient [14], [15]. Thus, it becomes a challenge to provide a better monitoring system that could be able to progressively monitor the outcome throughout the treatment.

We approach the problems by developing a computational method using Dynamic Joint-Sentiment-Topic Model (dJST) [16] to measure and monitor the patient treatment outcome by tracking patient's current and recurrent views of topic and sentiment. Specifically, by incorporating the sentiment and topic analysis for each therapy session, we contribute to the identification of sentiment and topic trend evolved throughout treatments on the author level. We highlight the sentiment (positive or negative) and the topic of each therapy session for each patient. We show the high potential of using this computational method in clinical settings by comparing our analysis with the professional practitioner analysis.

## II. MATERIALS AND METHODS

Sentiment-topic models have been successfully applied for a wide range of opinion mining tasks, such as aspect-based sentiment analysis [17], and contrastive opinion mining [18]. To discover and track the sentiment and topic over time from each therapy session, we employed the Dynamic Joint Sentiment Topic (dJST) [16] model. The model as shown in Fig. 1 assumes that the documents at current epoch are influenced by documents in the past, where the current sentiment-topic specific word distributions $\varphi_{l,z}^t$ at epoch $t$ are generated according to the word distributions at previous epochs. The time stamp $t$ for each stream of documents $\{d_1^t, \cdots, d_M^t\}$ can be an hour, a day, or a year. Each document d at epoch t is represented as a vector of word tokens, $w_d^t = \left(w_{d_1}^t, w_{d_2}^t, \cdots, w_{d_{N_d}}^t\right)$.

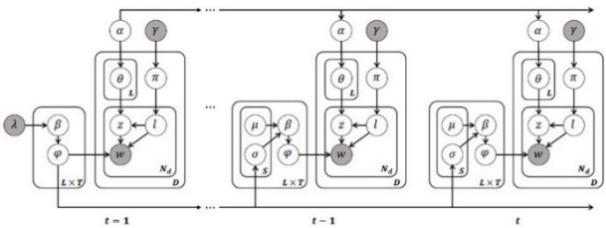

Fig 1. The Dynamic JST model.

The evolutionary matrix of topic $z$ and sentiment label $l$, $E_{l,z}^t$ where each column in the matrix is the word distribution of topic $z$ and sentiment label $l$, $\sigma_{l,z,s}^t$, generated for document streams received within the time slice specified by $s$, where $s \in \{t-S, t-S+1, \cdots, t-1\}$, the current sentiment-topic-word distributions are dependent on the previous sentiment-topic specific word distributions in the last $S$ epochs. We then attach a vector of $S$ weights $\mu_{l,z}^t = [\mu_{l,z,0}^t, \mu_{l,z,1}^t, \cdots, \mu_{l,z,S}^t]'$, each of which determines the contribution of time slice $s$ in computing the priors of $\varphi_{l,z}^t$. Hence, the Dirichlet prior for sentiment-topic-word distributions at epoch $t$ is $\beta_{l,z}^t = E_{l,z}^{t-1} \mu_{l,z}^t$.

Assuming we have already calculated the evolutionary parameters $\{E_{l,z}^{t-1}, \mu_{l,z}^t\}$ for the current epoch $t$, the generative story of dJST as shown in Fig. 1 at epoch $t$ is given as follows:
- For each sentiment label $l = 1, \cdots, L$
- For each topic $z = 1, \cdots, T$
  - Draw $\alpha_{l,z}^t | \alpha_{l,z}^{t-1} \sim \Gamma(\nu \alpha_{l,z}^{t-1}, \nu)$
  - Compute $\beta_{l,z}^t = \mu_{l,z}^t E_{l,z}^t$
  - Draw $\varphi_{l,z}^t \sim Dir(\beta_{l,z}^t)$
- For each document $d = 1, \cdots, D^t$
  - Choose a distribution $\pi_d^t \sim Dir(\gamma)$.
  - For each sentiment label $l$ under document d, choose a distribution $\theta_{d,l}^t \sim Dir(\sigma^t)$.
  - For each word $n = 1, \cdots, N_d$ in document $d$
    * Choose a sentiment label $l_n \sim Mult(\pi_d^t)$,
    * Choose a topic $z_n \sim Mult(\theta_{d,ln}^t)$,
    * Choose a word $w_n \sim Mult(\varphi_{ln,zn}^t)$

The Dirichlet priors $\beta$ of size $L$ x $T$ x $V$ are first initialised as symmetric priors of 0.01, and then modified by a transformation matrix $\lambda$ of size $L$ x $V$ where encodes the word prior sentiment information. $\lambda$ is first initialised with all the elements taking a value of 1. Then for each term $w \in 1, \cdots, V$ in the corpus vocabulary, the element $\lambda_{lw}$ is updated as follows:

$$\lambda_{lw} = \begin{cases} 0.9 & if\ f(w) = 1 \\ 0.05 & otherwise \end{cases}$$

where the function $f(w)$ returns the prior sentiment label of $w$ in a sentiment lexicon, i.e., positive, or negative.

### A. Experimental Setups

In this section, we describe the datasets of counselling transcripts from Carl Roger's therapy sessions, and discuss the settings used in our experiments.

#### 1) Dataset

We conducted the experiments using the transcripts of Carl Rogers' therapy sessions - the founder of client-centred psychotherapy. We acquired the collection of 158 transcripts from 51 clients of Roger's therapy cases from [19]. The cases reflect the wide range of clients with whom he worked from schizophrenic patients to a conflicted clinical psychologist. This dataset has been made available for research purposes [19]. The Rogers' psychotherapy transcripts have been used in psychotherapy research [20]–[25] and analysing the cause of depression [26]. For the sake of our model evaluation, we selected five cases of client session transcripts that have been analysed by the professional psychotherapist. We dropped the remaining 46 cases in this study. For each case, we only extracted the client's verbatim transcripts and do not consider the therapist verbatim in this study. We show a summary of clients and the number of sessions in TABLE I and summarised each case in the following paragraph.

TABLE I.
ROGERS TRANSCRIPTS DATASET STATISTICS.

| Client | No. of Session | Average doc length | Vocabulary Size |
|---|---|---|---|
| Frank | 5 | 131 | 384 |
| Bryan | 9 | 1068 | 2582 |
| Marry | 7 | 279 | 717 |
| Vib | 9 | 879 | 1980 |
| Int | 7 | 712 | 1625 |

i. *The case of Mr. Herbert Bryan:* Bryan, a client who was suffering from a blocking and a variety of neurotic complaints. He was suffering severe pain from his blocking, which interferes his sexual, business, life and social life [27].

ii. *The case of Frank:* Frank, a client who was a student in college having serious attitude problems. He has shown argumentative, attention-getting, and uncooperative behavior in his classroom [27].

iii. *The case of Marry Jane Tilden:* Marry Jane, a client who was a high school graduate from an upper-middle-class family. She was brought to hospital by her mother who worried as she noticed Marry had symptoms of major depression with the presence of suicidal ideation, social isolation, low self-esteem, and strong self-critical attitudes.

*2) Settings*

Each dataset underwent pre-processing including conversion to lowercase, removal of non-alphanumeric characters, and removal of stop words. We empirically set the number of topics to 5 for the 2 sentiment labels (i.e., positive and negative), which is equivalent to a total of 10 sentiment-topic clusters for each case.

TABLE II.
INDEX OF BEHAVIORAL MATURITY IN THE TEN CASES BY [28].

| Case | Interview Number | | | | | | | | |
|---|---|---|---|---|---|---|---|---|---|
| | 1 | 2 | 3 | 4 | 5 | 6 | 7 | 8 | 9 |
| Miss Ban | 1.3 | 1.3 | 3.3 | 1.0 | 1.7 | 1.0 | 1.2 | 2.7 | 1.8 |
| Mrs. Dem | 1.0 | 1.0 | 1.4 | | | | | | |
| Mr. Far | 2.3 | 1.8 | - | 2.0 | 2.3 | - | - | | |
| Miss Int | 1.5 | 1.9 | 1.4 | 1.0 | 4.0 | 4.0 | 2.5 | | |
| Mr. Que | 2.0 | 1.8 | 2.0 | 2.0 | 2.0 | 4.0 | - | | |
| Mrs. Sim | 2.3 | 1.3 | 1.0 | 1.5 | 2.0 | 2.3 | 2.3 | | |
| Mr. Sketch | - | 1.0 | 1.2 | | | | | | |
| Miss Vib | 1.0 | 1.0 | 1.8 | 1.5 | 1.1 | 1.0 | 1.0 | 1.6 | 3.8 |
| Miss Wab | 1.1 | - | 1.0 | | | | | | |
| Mr. Win | 1.6 | 2.4 | 3.7 | - | 3.6 | | | | |

III. RESULTS AND DISCUSSION

In this section, we present our results and analysis of the experimental datasets. We aimed to analyse the session by focusing on the sentiment and topic trend throughout psychotherapy session for each client. We first show the results from the two clients i.e., Miss Vib and Miss Int. For these two cases, however, we couldn't find any detailed analysis evaluated by the expert. Therefore, we only demonstrate the sentiment analysis from our model and compare with the results reported in the studies by [28]. We show the excerpt of the results from [28] in Table II. The table represents the index of behavioural maturity with a scale of 1 to 4, with defined values of 1, 2 and 4 towards the changes in counselling therapy. For value 1, it indicates the behaviour of little or no control over himself/herself or the environment. Value 2 indicates the individual has some control over the environment; whereas value 4 indicates good behaviour with self-direction, maturity and responsibility [28].

*A. Analysis of Sentiment-Topic for Miss Vib*

We show the trend of sentiment analysis over the therapy session of Miss Vib in Fig. 2. X-axis in the figure represents the therapy session; whereas Y-axis represents the probability of sentiment given a document $P(l|d)$. As depicted, negative sentiments are dominant throughout the course. However, it is observable that the positive sentiment value has significantly raised between session 8 and 9. It shows that our sentiment analysis trends are consistent with the reported behaviour trends analysed in the work of [28] as shown in Table II, in which towards the end of the session (session 9) the maturity level was increased.

Apart from the sentiment analysis, we also show some of the topics extracted from the Miss Vib counselling sessions in Table III. From the topic words e.g., '*family*', '*friends*', '*married*', we can suggest that for instance Topic 1-negative is related to the relationship; Topic 5-negative is more related to education based on the topic words such as '*courses*', '*library*' etc. Whereas, Topic 1-positive could be analysed as the topic related to life's goal from the common topic words such as '*aim*', '*goal*', '*accomplished*'. Similarly, we could suggest that from the topic words, Topic 2-positive is related to education/study based on the topic words such as '*study*', '*teachers*', and '*scholarship*'.

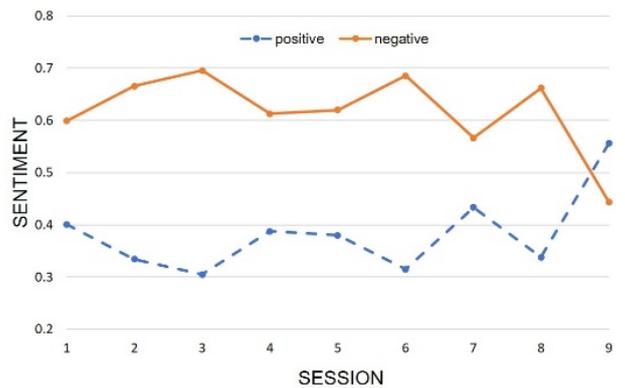

Fig 2. The Case of Miss Vib.

TABLE III.
THE CASE OF MISS VIB - EXAMPLES OF TOPIC.

| Sentiment | Topic | Examples of topic words |
|---|---|---|
| Negative | 1 | home family worried life finished felt sort friends upset wondering end write normal experience worry mother working studied married lost ... |
| Negative | 5 | feeling courses sort bit reading group morning study reason stay learned |

| | | material library possibilities problem kids appointment fright term gradually department ... |
| --- | --- | --- |
| Positive | 1 | year occurred children goal sort camp letting settled apparently lived smile women gave standards news circumstances satisfied conference aim accomplished ... |
| Positive | 2 | study lower physical children happened topic growth month interviews worry school beginning public field teachers state awful material research progress station scholarship relations class health fellowship ... |

*B. Analysis of Sentiment-Topic for Miss Int*

We show the trend analysis of sentiment over the therapy session of Miss Int in Fig.3. As depicted, there is no consistent trend evident for 7 counselling sessions. From the analysis, we observed that for the first three sessions, positive sentiment is relatively dominant over the negative sentiment. From session 3 onward, however, the negative sentiment significantly raised and stayed dominant towards the end. When compared to [28], we found no similarities with the level of maturity as shown in Table II. This contradiction analysis might be due to the differences of interpretation of sentiment words by our model.

Despite the sentiment analysis, we illustrate some of the topics extracted from the Miss Int counselling sessions in Table IV. Based on the examples of topic words, we can suggest that Miss Int expressed positive-sentiment towards topics related to the topic *education* and topic *life*. In contrast, there are also negative-sentiment topics expressed related to the topic of her life *expectation* and topic *living*. We acknowledge that it is quite challenging to learn the topic words without the background knowledge for each client.

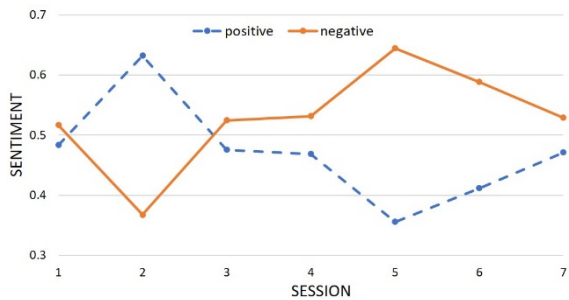

Fig 3.The Case of Miss Int.

TABLE IV.
THE CASE OF MISS INT - EXAMPLES OF TOPIC.

| Sentiment | Topic | Examples of topic words |
| --- | --- | --- |
| Negative | 1 | expectation understand long surprised solve girl big telling hard problems expectations neurotic hurting afraid shocked puts distance break scared altruism ... |
| Negative | 5 | time home felt grades job sitting made awful close point energy settle waste completely exist living ready social analyzing wasting worried ... |
| Positive | 1 | things feel people guess work lot thinking life make realize back figure study find business talking reason idea experience learning ... |
| Positive | 4 | future notes exam credit read present finals stay reading wait taking sit graduate cautious system meanings courses term happen grade ... |

*C. Analysis Evaluation*

In this section, we present the details of further analysis from our experimental results. Specifically, we compare our analysis finding with the overall commentary made by the professional psychotherapist published in [27], [29]. Due to the availability of published analysis by the domain experts, we only evaluate three of our analysis cases, i.e. Herbert, Frank, and Marry Jane Tilden.

*i. Analysis of the Case of Herbert Bryan*

We show the trend analysis of sentiment distributions over the therapy session of Bryan in Fig. 4. As depicted, the first five sessions showed high negative sentiments in Bryan's therapy. Nevertheless, Bryan has ended the therapy with positive sentiments, in which the attitude has started to change from session 6 onward. This finding is consistent with the concluding remarks made by Rogers for the case of Herbert Bryan [27]. Below is the excerpt by Rogers over Bryan's therapy sessions:

*"The extremely positive feelings and actions and the self-confidence expressed are in the most astonishing contrast to the first three interviews, or to Interviews Five and Six. He has completed the full cycle of therapy - expression, insight, positive decision, and reoriented action in line with the newly chosen goals."*

Also, it is worth mentioning that, as shown in Fig. 4, the negative sentiment value is constantly reduced from session 1 to session 4. That is, we observed an increment in the positive sentiment value. This positive change is confirmed by the analysis made by Rogers, as per below excerpt from the session 4 [27]:

*"A comparison of these attitudes toward the self with those expressed in previous interviews indicates clearly the tremendous development in insight and the increasingly positive attitudes."*

However, it is noticeable that the negative sentiment again significantly raised up on session 5. Rogers's analysis on Bryan's discouragement and de-motivation during this session could explain this trend. Below is the excerpt [27]**:**

*"The full measure of his discouragement, as he faces the implications of the insight achieved in the previous contact, is best shown by listing, as before, the spontaneous sentiments voiced during the interview. I haven't any motivation to choose the better way. When I have nothing but neurotic satisfactions, it is hard to feel that other satisfactions would be better. I feel discouraged about myself. I'm suffering real pain. I should like to have you pull a rabbit out of a hat for me. This whole struggle is very exhausting to me."*

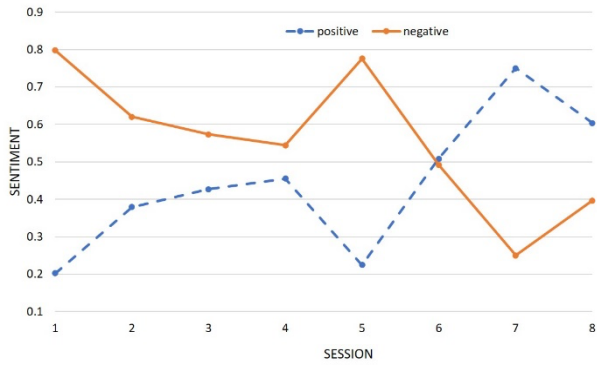

Fig 4. The Case of Herbert Bryan - Sentiment Distributions.

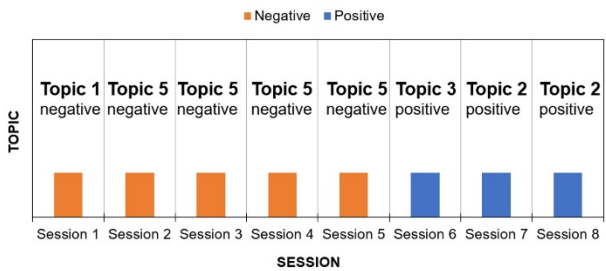

Fig 5. The Case of Herbert Bryan - Topic Distributions.

TABLE V.
THE CASE OF HERBERT BRYAN - EXAMPLES OF TOPIC.

| Sentiment | Topic | Examples of topic words |
|---|---|---|
| Negative | 1 | blocking blocked feel logic speech inhibition voyeurism impulse intercourse release energy superior physical difficulty intense ideological girl childhood abdomen psychological ... |
| Negative | 5 | feel feeling neurotic time thing change situation job life negative thought definite condition money painful sexual intellectual reason emotional sex ... |
| Positive | 3 | satisfactions faith means ahead act deep situations seated confidence radical behavior neurosis achievement environmental progress change knowledge cure growth experience ... |
| Positive | 2 | improvement technique job things happen night evening valuable public test react fluctuations social literature lead defense absolutely fantastical deeper efficiency... |

Apart from that, we also report the topic proportions for each therapy sessions in Fig. 5. Some examples of topics can be seen in Table V. From the result, we can generally suggest that Bryan has expressed his feeling and thoughts on Topic 1 in the first session, which is about the topic related to *blocking* symptoms that he has suffered from. This finding is consistent with the conclusion commentary made by Rogers as published in [27]. Below is the excerpt from the summary commentary by Rogers in session 1:

*"The following would seem to be a fair summary of the outstanding attitudes, which have been spontaneously expressed: I suffer from a blocking which interferes with my sexual life, my business life, my social life. I suffer excruciating pain from this blocking. My only satisfaction is voyeurism."*

As for the four subsequent sessions (session 2-5), Topic 5 has been consistently discussed in Bryan's session. As can be seen in Table V, the topic is related to the feeling of *neurotic*. This finding is consistent with the Rogers analysis. Below is the excerpt from the summary commentary by Rogers [27]:

*"The major feelings which have been spontaneously expressed. The following would seem to be the major attitudes: My neurosis is resisting treatment. I cannot carry through the jobs I have lined up. In my present condition, I cannot face all the difficulties. If I do work it is a terrific struggle and it leaves me exhausted. I obtain partial release through night clubs. I keep my love affairs only as long as I want. I liked one girl, but was afraid of the responsibility of marriage. I feel jealousy about my girlfriends, without reason. I feel that I overvalue sex at the present time. My parents are to blame for my lack of sexual satisfaction."*

As depicted in Fig. 4, from session 5 onwards, the negative sentiment has dropped significantly. Yet, Bryan has completed the therapy with the positive sentiment, in contrast to the earlier 5 sessions. For these last three sessions, Bryan has discussed on the Topic 2 and Topic 3, which is related to *satisfaction* and *improvement*. Again, this finding is confirmed by Rogers's analysis as shown in the excerpt below [27]:

*"These attitudes show very vividly the fact that, after teetering for two interviews between neuroticism and growth, Mr. Bryan has chosen the pathway of growth with a clearness and vitality that is amazing. Between the sixth and seventh interviews the accumulated insight has been translated into a positive decision, which brings a decided feeling of release. The attitudes expressed are in sharp contrast to the weakness and helplessness which were evident in the two preceding interviews. The crisis is fully passed. The client has discovered resources within himself for making this crucial choice and moving ahead."*

*ii. Analysis of the Case of Frank*

We show the trend analysis of sentiment distributions over five therapy sessions of Frank in Fig. 6. As depicted, Frank started the therapy session with high negative sentiment. The negative attitude has constantly dropped over the session, that is, the highest positive sentiment reached at session 3. However, the negative sentiment again raised until the end of the session. As can be seen, Frank has ended the session with high negative sentiment. However, when we compared the result with the analysis made by Rogers [27], we noticed that only the first four sessions are consistent with Rogers's analysis. Below is the excerpt by Rogers over Frank's therapy sessions:

*"Although the first two interviews are largely ``talking out'' processes there are a few statements which indicate the beginnings of self-understanding. In the third interview, particularly on pages 8 and 9, significant insight is achieved.... The fourth interview represents a distinct slump in progress... The fifth interview contains not only a fresh surge of insight but many new choices and positive actions."*

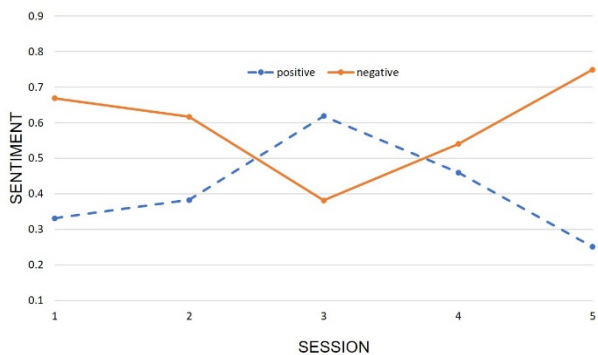

Fig 6.The Case of Frank - Sentiment Distributions.

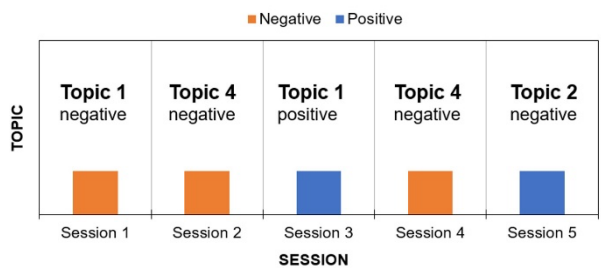

Fig 7.The Case of Frank - Topic Distributions.

TABLE VI.
THE CASE OF FRANK - EXAMPLES OF TOPIC.

| Sentiment | Topic | Examples of topic words |
|---|---|---|
| Negative | 1 | procrastinate goal college diagnose situations difficulty sort feel explain dean makes dorm kicking wasting trouble verbal defy digressions students unusual simple ... |
| Negative | 4 | friends problem sincere inside learn instance avoid eyes objective exploding seat dangerous trouble decision psychology limitations feelings jail emotional afraid ... |
| Negative | 2 | thinking work easier job confidence circle stand wonders pay places depot army troubles office retards shilly shallying rid blackboard grow ... |
| Positive | 1 | responsibility decisions exam acquire frankly changing workers education questioned highly decide background visualize statements acquaintance wishful retreats purposes rethink goal ... |

Also, we report the topic proportions for each therapy session in Fig. 7. We show some examples of the topic in Table VI. Based on the results, we can generally suggest that during Session 1 Frank mentioned topics related to *procrastinate*. Topic 4 related to *friends* has been discussed in Session 2 and 4. Whereas, in session 3, Frank has mentioned the topic related to *responsibility*. As to compare with Rogers analysis, for the case of Frank, only the analysis for session 1 and 5 are available. Below is the excerpt from Rogers's analysis:

*"Summary of Attitudes in First and Last Interviews. First Interview: I'm here because I'm sent. I am a little bothered by my procrastination. I have a bad past. I'm always in trouble. I am apathetic and lax about meeting responsibilities.... Fifth Interview: I met my responsibility for the psychological examination. I am meeting many situations responsibly. I expect to change. I've given up blaming my early experiences..."*

### iii. Analysis of the Case of Marry Jane Tilden

We show the trend analysis of sentiment distributions over seven therapy sessions of Marry Jane in Fig. 8. Note that the transcript for session 2, 4, 6, and 8 are not available. Thus, we skipped all those sessions for the analysis. As depicted, Marry started the therapy session with very high negative sentiment, and negative sentiment is steadily declined over the course. Nevertheless, the positive sentiment increased significantly between session 9 and 11. Marry had ended the session with a significantly high positive sentiment.

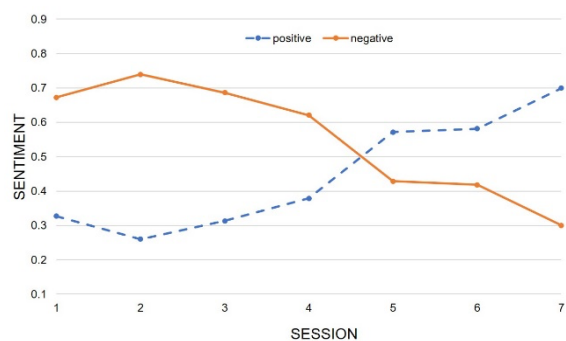

Fig 8.The Case of Marry Jane - Sentiment Distributions.

For Marry Jane session 1, our analysis showed that a high value of a negative sentiment is consistent with the analysis by Rogers as significant feelings were expressed in the first session. See below the excerpt [29]:

*"Significant feelings expressed in the first interview: Everything is wrong with i.e., I feel I'm abnormal.. I have lost faith in everything, especially myself..I have absolutely no self-confidence.."*

Marry continued her third session with negative sentiment, in which the sentiment values are higher than the previous session. However, we noticed that the positive sentiment is started to grow in this session. Likewise, negative sentiment values are reduced. Rogers analysis on Marry third session is reported in the below excerpt [29]:

*"The client expresses and idea very basic in her thinking, the question of whether she is capable or as intelligent as other people are.. Client discuss the very significant problem of motivation, both with regard to herself and to other people."*

By session 5, despite that there is a significant rise in positive sentiment, however, the negative sentiment remains high. Below is the excerpt from Rogers's analysis [29]:

*"..major attitudes expressed by Marry Jane. I'm afraid to venture, because i'm afraid i will fail. I was frightened by my jealousy when she went with a boy in whom i was interested. It made me afraid to trust my feelings."*

As can be seen in, the negative sentiment is dramatically declining between session 7 and 9; whereas the positive sentiment has significantly increased. Our result is confirmed by Roger's analysis as illustrated in the below excerpt [29]:

*"Marry Jane discusses her plan for work in a rather discourages way, fearful that "something will turn up and I'll sort of lose faith all over again..."*

From session 9 to 11, the trend shows the major changes in Marry's therapy session. That is, the positive sentiment is dominant as there are further declines in the negative sentiment. Our trend analysis is consistent with Rogers, as per below excerpt for session 9, 10, 11 respectively:

*"She voices the attitude that she is not too well connected with reality. She discusses this and realizes that many of her past problems lies within herself....she realizes that she is always dissatisfied with anything she does-her job or any other undertaking.."*

*"She continues with a discussion of the fact that when she is feeling good she feels much more adventorous, and is now thinking of taking a very different and interesting job during the coming summer, with a girl with whom she has become friendly at work... She tells f a friend she mas made, a very interesting girl... She continues to discuss the family situation. .."*

*"Significant feelings expressed in the eleventh interview. I am really changing.I am taking a new course of my own choosing. I realise i am actually like other people. I feel more free... I can accept marriage...I've been out much more with friends and enjoy it...."*

| Client | Our Analysis | | | | | | | | | Domain Expert Analysis | | | | | | | | |
|---|---|---|---|---|---|---|---|---|---|---|---|---|---|---|---|---|---|---|
| | S1 | S2 | S3 | S4 | S5 | S6 | S7 | S8 | S9 | S1 | S2 | S3 | S4 | S5 | S6 | S7 | S8 | S9 |
| The case of Miss Vib | N | N | N | N | N | N | N | N | P | N | N | N | N | N | N | N | N | P |
| The case of Miss Int | N | P | N | N | N | N | - | - | - | N | P | N | N | N | N | - | - | - |
| The case of Bryan | N | N | N | N | P | P | P | P | P | N | N | N | N | P | P | P | P | P |
| The case of Frank | N | N | P | N | N | - | - | - | - | N | N | P | N | P | - | - | - | - |
| The case of Marry | N | N | N | P | P | P | - | - | - | N | N | N | P | P | P | - | - | - |

Fig 9. The Summary of Analysis between Our Method and Expert Analysis. Note: N represents negative sentiment; P represents positive sentiment; - represents not available

### D. Summary of Analysis Evaluation

We show the summary of analysis evaluation between our computational method and expert analysis in Fig. 9. We compare the analysis from domain expert with our proposed computational method analysis. As shown in the figure, it clearly indicates that our method illustrates comparable results with an expert's analysis. Our results present a similar analysis with the domain expert. For the case of Miss Vib, Miss Int, Bryan and Marry, we achieve 100% accuracy. However, in session 5 (S5) of Frank's case, our analysis is contradicted with the domain expert analysis, in which we analyse negative sentiment (N) instead of positive (P).

## IV. CONCLUSION

In this paper, we have demonstrated the work on monitoring the progress of outcome from the individual psychotherapy counselling transcripts. Progress monitoring is important to assess the effects of changes on knowledge, attitudes, beliefs, and behaviours upon the psychotherapy intervention. By employing the topic model for monitoring the progress of counselling sessions, it provides the trend of sentiments and topics evolve over the course of treatment for each client. We evaluated the method by comparing the analysis from our model with the analysis made by the domain expert.

From the experimental findings, our model revealed comparable results with the analysis made by the domain expert. Therefore, we suggest that the dynamic topic models provide an opportunity for tracking the progress of the client by closely analysing the shifts of sentiment and topic for each session. This monitoring might be beneficial to serve as an early identification tool in analysing whether the necessary change in therapy needs to be taken.

We believe that with further enhancement in our model, for instance by incorporating more domain-specific lexicons as utilised in [30], our result could be further improved. For future work, we could also integrate the counsellor verbatim transcripts for the analysis. Apart from that, we also observed that it is difficult to predict the future outcome of the next counselling session. As we noticed that the number of counselling sessions is not linked to the changes of positive attitudes towards the treatment, as no evidence of trend can be used as a guide to predict sentiment outcome. We suggest that this might be different for each case and depends on the client's characteristics and the complexity of clients' problems.

ACKNOWLEDGMENT

This paper has been supported by Center for Advanced Computing Technology (C-ACT), Fakulti Teknologi Maklumat dan Komunikasi (FTMK), Universiti Teknikal Malaysia Melaka, Malaysia.

Treatment for Depression," *Arch Gen Psychiatry*, vol. 61, no. 7, pp. 714--719, 2004.